\documentclass[preprint,12pt,authoryear]{elsarticle}

\usepackage{amssymb}
\usepackage{amsmath}

\usepackage{wrapfig}
\usepackage{hyperref}

\usepackage{multirow}
\usepackage{soul}
\usepackage{url}
\usepackage[utf8]{inputenc}
\usepackage{graphicx}
\usepackage{amsmath}
\usepackage{amsthm}
\usepackage{booktabs}
\usepackage{algorithm}
\usepackage{algorithmic}
\usepackage[switch]{lineno}

\usepackage{balance}
\usepackage{algorithmic}
\usepackage{algorithm}
\usepackage{array}
\usepackage{newtxmath}
\usepackage{textcomp}
\usepackage{stfloats}
\usepackage{url}
\usepackage{verbatim}
\usepackage{graphicx}
\usepackage{colortbl}
\usepackage{diagbox}

\def\ie{{\em i.e.}}
\def\eg{{\em e.g.}}

\newcommand{\bl}[1]{\textbf{#1}}

\newcommand{\mc}[1]{\mathcal{#1}}

\newcommand{\myPara}[1]{\vspace{.05in}\noindent\textbf{#1}}

\journal{Neural Networks}

\begin{document}

\begin{frontmatter}



\title{PKI: Prior Knowledge-Infused Neural Network for Few-Shot Class-Incremental Learning} 




\author[label1,label2]{Kexin Bao}
\ead{baokexin@iie.ac.cn}
\author[label3]{Fanzhao Lin}
\ead{linfanzhao@sgepri.sgcc.com.cn}
\author[label4]{Zichen Wang}
\ead{zichen\_wang@gwmail.gwu.edu}
\author[label1,label2]{Yong Li}
\ead{liyong@iie.ac.cn}
\author[label5]{Dan~Zeng}
\ead{dzeng@shu.edu.cn}
\author[label1,label2]{Shiming~Ge\corref{cor1}}
\ead{geshiming@iie.ac.cn}
\cortext[cor1]{Shiming Ge is the corresponding author.}

\affiliation[label1]{organization={Institute of Information Engineering, Chinese Academy of Sciences},
            city={Beijing},
            postcode={ 100092}, 
            country={China}}

\affiliation[label2]{organization={School of Cyber Security at University of Chinese Academy of Sciences},
            city={Beijing},
            postcode={100049},
            country={China}}

\affiliation[label3]{organization={NARI Technology Company Limited},
            city={Nanjing},
            postcode={210000},
            state={Jiangsu},
            country={China}}

\affiliation[label4]{organization={School of Engineering and Applied Science, the George Washington University},
            city={Washington},
            postcode={DC 20052},
            country={USA}}
\affiliation[label5]{organization={Department of Communication Engineering, Shanghai University},
            city={Shanghai},
            postcode={200040},
            country={China}}




\begin{abstract}
Few-shot class-incremental learning (FSCIL) aims to continually adapt a model on a limited number of new-class examples, facing two well-known challenges: catastrophic forgetting and overfitting to new classes. Existing methods tend to freeze more parts of network components and finetune others with an extra memory during incremental sessions. These methods emphasize preserving prior knowledge to ensure proficiency in recognizing old classes, thereby mitigating catastrophic forgetting. Meanwhile, constraining fewer parameters can help in overcoming overfitting with the assistance of prior knowledge. Following previous methods, we retain more prior knowledge and propose a prior knowledge-infused neural network (PKI) to facilitate FSCIL. PKI consists of a backbone, an ensemble of projectors, a classifier, and an extra memory. In each incremental session, we build a new projector and add it to the ensemble. Subsequently, we finetune the new projector and the classifier jointly with other frozen network components, ensuring the rich prior knowledge is utilized effectively. By cascading projectors, PKI integrates prior knowledge accumulated from previous sessions and learns new knowledge flexibly, which helps to recognize old classes and efficiently learn new classes. Further, to reduce the resource consumption associated with keeping many projectors, we design two variants of the prior knowledge-infused neural network (PKIV-1 and PKIV-2) to trade off a balance between resource consumption and performance by reducing the number of projectors. Extensive experiments on three popular benchmarks demonstrate that our approach outperforms state-of-the-art methods.

\end{abstract}




\end{frontmatter}




\section{Introduction}

In many real-world scenarios, data is received continuously with new classes emerging regularly \citep{DBLP:journals/csur/GomesBEB17}. Class-incremental learning (CIL) incrementally expands their knowledge and discriminates new classes, where the primary challenge lies in preventing catastrophic forgetting~\citep{DBLP:conf/eccv/ChaudhryDAT18}. Given the high costs of collecting and labeling data, new classes may only have a few labeled data available. In this situation, CIL methods struggle in scenarios with limited labeled data. And few-shot class-incremental learning (FSCIL) is proposed to incrementally acquire new knowledge from limited examples while retaining previous knowledge~\citep{tao2020cvpr,achituve2021icml,ahmad2022cvprw,yang2023iclr}, which is dedicated to solving catastrophic forgetting and overfitting. Catastrophic forgetting leads to a decline in performance in old classes due to losing previously learned concepts when adapting to new classes. Overfitting is a risk that the model performs well on the training set but poorly on the test set. These two challenges pose significant obstacles for FSCIL, as it strikes a delicate balance between accommodating new knowledge and efficiently retaining previously learned knowledge.



\begin{figure}[th]
	\centering
	\includegraphics[width=1.0\linewidth]{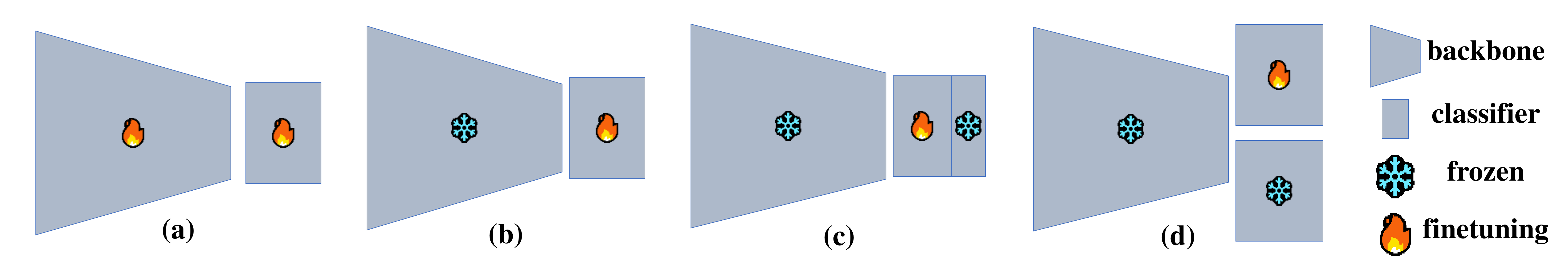}
	\caption{The network for FSCIL can be separated into {{two major components (\eg, a backbone and a classifier).}} Existing methods (a-c) tend to freeze more components or parameters and make relatively fewer adjustments in the incremental phase. Inspired by that, our approach (d) ensembles {{more preceding components}} to retain prior knowledge, effectively synchronizing old-class memorizing and new-class fitting.}
	\label{fig:motivation}
\end{figure}

To address the challenges, some early FSCIL methods regular the feature space by topology preserving and knowledge distillation during incremental sessions as shown in Figure~\ref{fig:motivation}(a)~\citep{tao2020cvpr,9645290}. However, finetuning numerous variable parameters with constraints preserves less prior knowledge, which makes the model quickly dilute old knowledge and overfitting. Recent methods {{emphasize saving an external memory and freezing the backbone as prior knowledge}} while regularizing the feature space during incremental sessions as shown in Figure~\ref{fig:motivation}(b)~\citep{chen2021iclr,xu2023kbs,chi2022cvpr}. The memory stores class-mean features of old classes to recall previous knowledge, ensuring continuity and stability across sessions. Freezing the backbone further ensures accurate preservation of prior knowledge to reduce the dilution of old knowledge, which is the foundation for current and subsequent sessions. However, adjusting the learnable classifier can lead to misalignment between the features and classifiers of old classes. Based on this, some methods further pre-assign a fixed {{matrix}} for final classification in the base session and only finetune other variable parts of the network in incremental sessions as shown in Figure~\ref{fig:motivation}(c)~\citep{peng2022eccv,yang2023iclr,song2023cvpr,ahmed2024cvpr}. {{In addition to the frozen backbone and the memory, the fixed matrix also serves as prior knowledge for incremental sessions}}, which transforms the challenge of misalignment between the features and classifier of old classes into feature alignment across different sessions. However, these methods often struggle to effectively push new-class examples into their corresponding feature spaces of old classes during training, as they primarily interpret new knowledge from the perspective of old knowledge. {{This limitation is particularly pronounced when dealing with limited examples.}} Recent methods tend to freeze more network components and {{store features in the external memory as prior knowledge}} for subsequent tasks. {{This way accelerates the memorization of old classes and}} reduces the risk of overfitting by minimizing parameter updates and applying regularization constraints. Based on these insights, we argue that the \textbf{prior knowledge from previous sessions is critically important for current and subsequent sessions {{to mitigate catastrophic forgetting and overfitting}} under FSCIL settings}. Following previous works, our research focuses on {{retaining and leveraging prior knowledge}}.

In this work, 
we propose a prior knowledge-infused neural network (PKI) to facilitate FSCIL. As shown in Figure~\ref{fig:motivation}(d), our main idea is to keep the parameters of the previously learned projectors to retain old knowledge and incrementally finetune a new projector for fitting novel classes. Following a backbone, the PKI cascades an ensemble of projectors, an extra memory, and uses a learnable fully connected layer as a classifier. In the base session, we train the whole model with sufficient data. After training, we freeze the backbone and the current projector, then calculate class-mean features to the memory. In each incremental session, we add a new projector to the ensemble for acquiring new concepts. Subsequently, we jointly train the current projector and the classifier with the assistance of other network components. After fine-tuning, we freeze the current projector and update the memory accordingly. The projector ensemble retains prior knowledge and learns knowledge continually. In this way, the prior knowledge is stored in the backbone, ensemble, and memory and is used as a guide for the classifier. PKI integrates prior knowledge from previous sessions and learns new knowledge flexibly, leading to a good trade-off between feature alignment of old classes and fitting over novel classes. Furthermore, given the overhead of cascading many projectors, we propose two variants of the prior knowledge-infused neural network (PKIV-1, PKIV-2) by shrinking the number of projectors, ensuring a minimal drop in performance and minimal resource overhead.



Our main contributions are threefold. First, we propose a prior knowledge-infused neural network to facilitate FSCIL by cascading an ensemble of projectors, which can effectively mitigate catastrophic forgetting and overfitting. Second, we propose two variants of the prior knowledge-infused neural network to reduce resource consumption and achieve a balance between performance and resource consumption. Third, we conduct extensive experiments and comprehensive comparisons on three popular benchmarks to demonstrate the effectiveness of our approach.

\section{Related Work}

\subsection{Catastrophic Forgetting} 
Catastrophic forgetting \citep{goodfellow2014iclr, kirkpatrick2017pnas,shi2021nips} is a ubiquitous phenomenon and remains a major capability challenge in deep learning. The popular methods for mitigating catastrophic forgetting are mainly based on structural or functional regularization. 
The structural regularization approach directly punishes the change of model parameters~\citep{kirkpatrick2017pnas,lee2020cvpr,10083158,10214591,10148995}. \citet{10214591} develops a self-paced regularization to reflect the priorities of past tasks and selectively maintain the knowledge amongst more difficult past tasks. \citet{10148995} preserves the topological structure of the feature space during the subsequent tasks. By contrast, the functional regularization approach stores and replays earlier data with an extra memory~\citep{rolnick2019nips,aljundi2018eccv,9849074,9939005,10335718}. \citet{9939005} exploit the sample with the smallest loss value to representatively characterize the corresponding class. \citet{lin2023supervised} incorporates information into knowledge distillation. GAN-like methods~\citep{wu2018nips,ostapenko2019cvpr} generate synthetic examples to align features. \citet{10335718} captures the inter-class diversity and intra-class identity. A good method to alleviate catastrophic forgetting often combines the advantages of both structural regularization and functional regularization.

\subsection{Class-Incremental Learning}

Class-incremental learning (CIL) focuses on incorporating the knowledge of new classes incrementally in the real-world~\citep{DBLP:conf/cvpr/GaoHDCWG23,DBLP:journals/nn/AnYZRWLH24,DBLP:conf/aaai/KimC21}. Existing CIL methods can be divided into three groups: replay of old classes, regularization, and architecture consolidation.

Replay of old classes maintains a small subset of old samples instead of discarding them. \citet{DBLP:conf/iccv/BelouadahP19} directly uses previous data of old classes for subsequent training. \citet{DBLP:journals/tnn/HoLDGX24} does not reserve real data of old classes but reserves features of old classes. Regularization regularizes network parameters to constrain changes in model parameters. \citet{DBLP:conf/cvpr/DharSPWC19} presents an information preserving penalty based on knowledge distillation. \citet{DBLP:conf/aaai/0002C21} constrains the parameter by means of feature, coverage maximization, and influence maximization. Architecture consolidation dynamically adjust the network structure during training. \citet{DBLP:conf/cvpr/MallyaL18} sequentially "pack" multiple tasks into a single network by performing iterative pruning and network re-training. \citet{DBLP:conf/eccv/0012PSJLT20} integrates an ensemble of auxiliary classifiers to estimate more effective regularization constraints.

\subsection{Few-Shot Class-Incremental Learning} 
Few-shot class-incremental learning (FSCIL) fulfills incremental learning with insufficient novel data~\citep{tao2020cvpr,achituve2021icml,ahmad2022cvprw}. It needs to make a trade-off between memorizing old classes and fitting novel classes~\citep{zhao2024tpami,snell2017nips} by adjusting the network. The network for FSCIL usually can be separated into a backbone, a projector and a classifier. Thus, many FSCIL methods are proposed with different degrees of network adjustment. 

Early methods update full models by designing networks~\citep{tao2020cvpr,zhao2024tpami} or distilling knowledge into new networks~\citep{cheraghian2021cvpr,zhao2023cvpr,cui2023tmm}. In general, these methods require a complex updating scheme for each incremental task and old knowledge always quickly dilutes because finetuning numerous variable parameters with constraints preserves less prior knowledge.

Current mainstream methods widely employ constraint finetuning schemes combined with prior knowledge from previous sessions~\citep{tian2024nn}, {which refers to the frozen backbone and an extra memory}~\citep{shi2021nips,zhang2021cvpr,zhu2021cvpr}. The memory constraint methods {{focus on the memory content}}, using a small buffer stored previous data to retain the prior knowledge~\citep{yang2023iclr,ji2023tip,chen2021iclr,zhang2021cvpr,zhu2021cvpr}. \citet{zhang2021cvpr} optimizes the graph parameters by sampling data from the previous datasets as the memory. \citet{yang2023iclr} adds more reference vectors calculated by real features to the memory for learning new tasks sequentially.  \citet{ji2023tip} ensembles multiple models for different memorized knowledge.
The weight regularization methods focus on constraint weights to identify old and new classes, minimizing the weight impact with the assistance of prior knowledge~\citep{10168925,10382651,yang2023tpami,akyurek2022iclr,hersche2022cvpr}. \citet{zhao2024tpami} designs a slow and fast network regularized by a frequency-aware strategy and develops a new feature space composition operation to enhance the inter-space learning performance. \citet{xu2023kbs} uses the base class feature space as the major and updates multiple new feature spaces as supplements based on prior knowledge. \citet{yang2023tpami} supports feature space via an adaptively updating network with compressive node expansion. These methods may suffer from misalignment between features and the variable network due to the adjustment of parameters, which brings catastrophic forgetting.
Based on previous mainstream methods, several methods pre-assign a fixed metric subspace for classification in the base session, {{which can be prior knowledge of incremental sessions to eliminate the uncertainty of classification subspace}}~\citep{pernici2020icpr,peng2022eccv,zhou2022cvpr,yang2023iclr,song2023cvpr,khandelwal2023masil,ahmed2024cvpr}.
\citet{yang2023iclr} holds a class average feature memory to keep the old knowledge and incrementally finetunes a projector with the assistance of the memory to achieve feature-classifier alignment. \citet{ahmed2024cvpr} {{builds a fixed orthogonal space for classification}}, then employs a combination of supervised and self-supervised contrastive loss based on the orthogonality in the representation space to improve the generalization of the embedding space. {{{These methods constraint parameters based on prior knowledge from previous sessions (\eg, the frozen backbone, memory, and pre-assigned fixed networks)}}, thereby playing an indispensable role. Prior knowledge helps alleviate catastrophic forgetting and provides additional regularization for subsequent sessions.}

\section{Approach}
\subsection{The Formulation of FSCIL}
In few-shot class-incremental learning (FSCIL)~\citep{tao2020cvpr,tian2024nn}, there is a stream of labeled training sets in time sequence $\mathcal{D}=\{\mc{D}^{(t)}\}_{t=0}^{T}$ with $\mc{D}^{(t)}=\{(\bl{x}_i,y_i)\}_{i=1}^{|\mc{D}^{(t)}|}$. Here, $\bl{x}_i$ is an example (\eg, image), $y_i\in L^{(t)}$ denotes its label where $L^{(t)}$ is the label space of the $t$-th training set. Any two class sets are disjoint, meaning $L^{(i)}\cap L^{(j)}=\varnothing$ for $\forall i, j$. 
Among them, $\mc{D}^{(0)}$ is a training set of the base session, and $\mc{D}^{(t)} (t>0)$ is the few-shot training set of novel classes in the $t$-th incremental session. In general, for $\mc{D}^{(t)} (t>0)$, the learning task is described by the \emph{$N$-way $K$-shot} setting where contains $N$ classes and $K$ examples per class. The task of FSCIL is learning a model $\phi(\bl{x};\bl{w})$ with parameters $\bl{w}$ that incrementally trains on $\mc{D}^{(t)}$ in the $t$-th session. After training, the model $\phi(\bl{x};\bl{w})$ is tested to recognize the classes in $\cup_{i=0}^{t} L^{(i)}$. {{The model $\phi=\{\phi_b,\phi_p,\phi_c\}$ consists of a backbone $\phi_b(\bl{x};\bl{w}_b)$ with parameters $\bl{w}_b$, a projector $\phi_p(\bl{f};\bl{w}_p)$ with parameters $\bl{w}_p$, and a classifier $\phi_c(\bl{v};\bl{w}_c)$ with parameters $\bl{w}_c$, where $\bl{w}=\bl{w}_b\cup\bl{w}_p\cup\bl{w}_c$ are the whole model parameters, $\bl{f}=\phi_b(\bl{x};\bl{w}_b)$ is the intermediate feature, and $\bl{v}=\phi_p(\bl{f};\bl{w}_p)$ is the vector. Besides, we also use a memory $\mathcal{M}$ in incremental sessions, which stores old knowledge to assist the model finetuning.}}
Thus, the FSCIL task is formulated into base learning $t=0$ (1a) and incremental learning $t>0$ (1b):
\begin{subequations}\label{eq:problem}
\begin{align}
\{\bl{w}_b,\bl{w}_p^{(0)},\bl{w}_c^{(0)}\}&=\arg\min_{\{\bl{w}_b,\bl{w}_c^{(0)}\}}\mathbb{E}_B(\mathcal{D}^{(0)};{{\bl{w}_b}},\bl{w}_p^{(0)},\bl{w}_c^{(0)}), \\
\{\bl{w}_p^{(t)},\bl{w}_c^{(t)},\mc{M}^{(t+1)}\}&=\arg\min_{\bl{w}_c^{(t)}}\mathbb{E}_I(\mathcal{D}^{(t)},{{\mc{M}^{(t)}}};\bl{w}_b,\bl{w}_p^{(t)},\bl{w}_c^{(t)}),
\end{align}
\end{subequations}
While training, the backbone is trained in the base session and frozen in incremental sessions.

\begin{figure*}[t]
\centering
\includegraphics[width=0.9\linewidth]{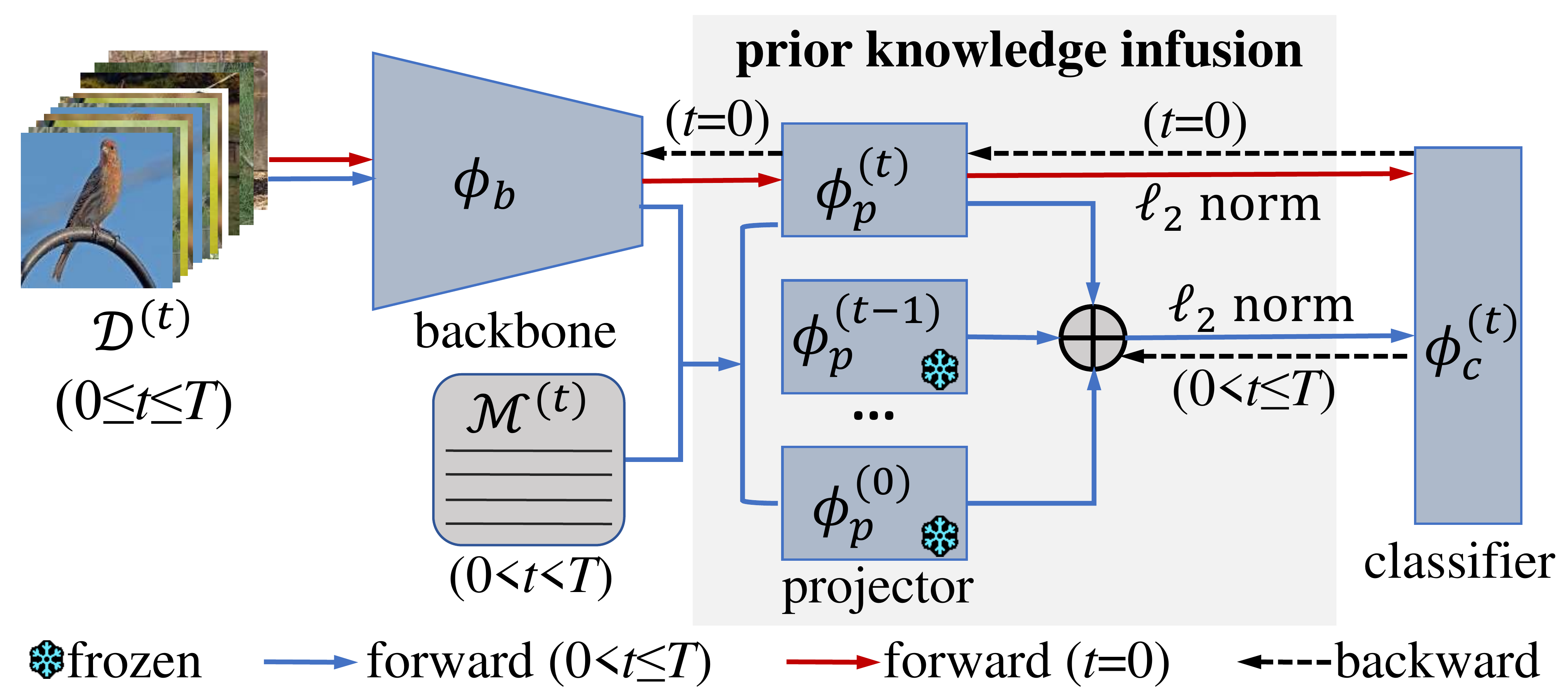}
\caption{
Our prior knowledge-infused neural network (PKI) is a two-phase approach for FSCIL. After a backbone $\phi_b(\bl{x};\bl{w}_b)$, a memory $\mc{M}^{(t)}$, a projector ensemble $\{\phi_p^{(0)},...,\phi_p^{(t)}\}$ and a classifier $\phi_c(\bl{x};\bl{w}_c)$ are incrementally updated in each session. 
In the base learning $(t=0)$, our approach trains the whole model on $\mc{D}^{(0)}$. In the incremental learning $(t>0)$, our approach freezes the backbone and incorporates the prior knowledge injection to jointly train the projector $\phi_p^{(t)}$ and the classifier $\phi_c^{(t)}$ on $\mc{D}^{(t)}$ under the supervision of previously learned $\mc{M}^{(t-1)}$ and $\{\phi_p^{(0)},...,\phi_p^{(t-1)}\}$.}
\label{fig:framework}
\end{figure*}

\subsection{Prior Knowledge-Infused Neural Network}

As shown in Figure~\ref{fig:framework}, our PKI consists of four components: a backbone  $\phi_b(\bl{x};\bl{w}_b)$ for feature extraction, an ensemble of projectors $\{\phi_p(\bl{f};\bl{w}_p^{(i)})\}_{i=0}^t$ for knowledge collection, a learnable fully-connected layer $\phi_c(\bl{v};\bl{w}_c)$ as a classifier, and a memory $\mc{M}^{(t)}$ storing the class means of intermediate features like~\citep{yang2023iclr,qi2018cvpr,agarwal2022mm}. Among them, prior knowledge exists in the whole model.

\myPara{Memory creation and update.} The extra memory $\mc{M}^{(t)}$ stores the class means of intermediate features in the preceding sessions, thereby the old-class knowledge could be efficiently remembered. To this end, the memory is created after the base learning and updated after each session:
\begin{equation}
\mc{M}^{{{(t+1)}}}=\{\mc{C}^{(0)},...,\mc{C}^{(t)}\},
\label{eq:memory}
\end{equation}
where $\mc{C}^{(i)} = \{{{(\bl{m}_{c},y_c) }}|~ c \in  L^{(i)}\}$ is the class means in session $i\in[0,t]$, and $\bl{m}_{c}=\frac{1}{|\mc{D}^{(i)}(y_i=c)|}\sum\nolimits_{\mc{D}^{(i)}(y_i=c)}\bl{f}_i$.

\myPara{Base learning.} Here, the ensemble of projectors only has one projector for model training. For each training example $\bl{x}_i$ on dataset $\mc{D}^{(0)}$, the backbone represents it into an intermediate feature $\bl{f}_i$. Subsequently, the feature $\bl{f}_i$ is projected by the projector into a vector $\bl{v}_i$, and an $l_2$ normalization is performed on $\bl{v}_i$ to get the output vector $\hat{\bl{v}}_i$. The vector $\hat{\bl{v}}_i$ is sent to the classifier to get the predicted label. Here, the vector $\hat{\bl{v}}_i$ involves the model parameters to be optimized:
\begin{equation}
\begin{aligned}
\bl{v}_i&=\phi_p(\phi_b(\bl{x};\bl{w}_b);\bl{w}_p^{(0)}), \hat{\bf{v}}_i = \bf{v}_i/|\bf{v}_i|.
\end{aligned}
\label{eq:output-feature}
\end{equation}
And in the base session, the backbone $\phi_b$, the projector {{$\phi_p^{(0)}$}} and the classifier {{$\phi_c^{(0)}$}} are jointly trained on $\mc{D}^{(0)}$ by minimizing the empirical risk loss:
\begin{equation}
\begin{aligned}
\mathbb{E}_B&=\sum_{(\bl{x}_i,y_i)\in\mc{D}^{(0)}}\mc{L}_{ce}(\hat{\bl{v}}_i,y_i),
\end{aligned}
\label{eq:baseloss}
\end{equation}
where $\mc{L}_{ce}$ is the cross-entropy loss.

After training, we freeze the backbone and the learned projector $\phi_p^{(0)}$. We calculate the class means of the intermediate features on training examples by Eq.~\eqref{eq:memory} to initialize the memory $\mc{M}^{(1)}$. 



\myPara{Incremental learning.} In each incremental session $(t>0)$, we add a new {{randomly initialized}} projector $\phi_p^{(t)}$ into the ensemble of projectors. The intermediate feature $\bl{f}_i$ of an example $\bl{x}_i$ on dataset $\mc{D}^{(t)}$ and {{the class mean $\bl{m}_c$ on the memory $\mathcal{M}^{(t)}$ are}} fed into the projector ensemble, and the projected features are subsequently aggregated to obtain the outputs as: 
\begin{equation}
\begin{aligned}
\bl{v}_i = \sum_{j=0}^{t}\phi_{p}(\bl{f}_i;\bl{w}_p^{(j)}), {{\bl{v}_c = \sum_{j=0}^{t}\phi_{p}(\bl{m}_c;\bl{w}_p^{(j)}).}}
\end{aligned}
\label{eq:projection_incremental}
\end{equation}
After {{normalizing the vector $\bl{v}_i$ and $\bl{v}_c$}}, we minimize the incremental empirical risk loss:
\begin{equation}
\begin{aligned}
\mathbb{E}_I=\sum_{(\bl{x}_i,y_i)\in\mc{D}^{(t)}}\mc{L}_{ce}(\hat{\bl{v}}_i,y_i)+\sum_{{{(\bl{m}_{c}}},y_{{c}})\in\mc{M}^{(t)}}\mc{L}_{ce}(\hat{\bl{v}}_{{c}},y_{{c}}).
\end{aligned}
\label{eq:incremental-loss}
\end{equation}
In Eq.~\eqref{eq:incremental-loss}, both the incremental examples $\mc{D}^{(t)}$ and the extra memory $\mc{M}^{(t)}$ are dedicated to constraining the projector $\phi_{p}^{(t)}$. Moreover, the previously learned projectors, \ie, $\{\phi_p^{(0)}, \phi_p^{(1)}, ..., \phi_p^{(t-1)}\}$ are fused into the learning, which implicitly provides effective supervision to optimize the output features of examples and aggregates the prior knowledge. 

After training, similar to the base session, we calculate the class means of the intermediate features on training examples by Eq.~\eqref{eq:memory} to update the memory $\mc{M}^{(t+1)} (0<t<T)$ and freeze the learned projector $\phi_p^{(t)}$. 

\myPara{Discussion.}~Different from existing methods, our PKI applies an incremental ensemble of projectors, which store more prior knowledge. In our PKI, sufficient prior knowledge exists in some components {{(the frozen backbone, frozen projectors, and memory), which improves the model to maintain stability and sustainability. Prior knowledge directly helps identify old classes by aligning output features with the ensemble, while indirectly facilitating the discrimination of new classes by accurately keeping old knowledge.}} Meanwhile, only updating a few parameters can reduce overfitting to new classes. Besides, we also find that the learning delivers a satisfactory accuracy after a small number of epochs, implying that our PKI can accelerate the model training. However, it needs more storage for the projector weights of all sessions. Further, we design two PKI variants in Section 3.3, ensuring a minimal drop in performance and minimal storage overhead.





\subsection{Variants of Prior Knowledge-Infused Neural Network} \label{variant}

Compared with other methods, PKI builds a new projector for each session and maintains an ensemble of projectors, which consumes many memory resources. We expect our method to achieve high accuracy while consuming as little memory resources as possible, promoting FSCIL quickly and efficiently. Based on this, we design two variants of the prior knowledge-infused neural network (PKIV-1 and PKIV-2) to trade off a balance between resource consumption and performance. Both of these variants focus on contracting the ensemble of projectors in each incremental session.



\myPara{PKIV-1.} In the whole training process, PKIV-1 only maintains one projector for learning knowledge. 
Before model training in each session, we {{randomly initialize}} the current projector weight $\bl{w}_p^{(t)}$, and add it to the previous projector weights. During model training in each session, we take examples on dataset $\mc{D}^{(t)}$ into the backbone and gain intermediate features $\{\bl{f}_i\}^{|\mc{D}^{(t)}|}$. Then, the memory $\mc{M}^{(t)}$ and the features are sent to the current projector. Instead of building a new projector for each session, PKIV-1 adds up the weights of the previous projectors and computes the output $\bl{v}_i$ as
\begin{equation}
\begin{aligned}
\bl{v}_i = \phi_{p}(\bl{f}_i;\sum_{j=0}^{t}\bl{w}_p^{(j)}).
\end{aligned}
\label{eq:pkiv-1}
\end{equation}
In Eq.~\eqref{eq:pkiv-1}, we only need to keep the prior knowledge $\sum_{j=0}^{t} \bl{w}_p^{(j)}$ after each session, which greatly reduces memory consumption. However, with the increasing of sessions, prior knowledge occupies a larger proportion in calculating outputs, which makes the model more difficult to finetune and learn new knowledge. Therefore, we design PKIV-2 to balance prior knowledge and new knowledge.

\begin{wrapfigure}[10]{r}{0.52\linewidth} 
\centering
\includegraphics[width=0.85\linewidth]{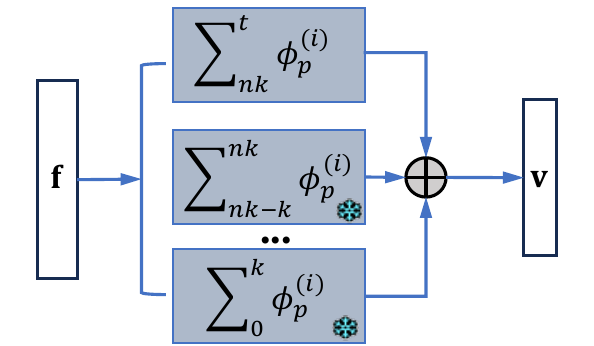}
\caption{Prior knowledge infusion of PKIV-2.}
\label{fig:vframework}
\end{wrapfigure}

\myPara{PKIV-2.} To balance prior knowledge and new knowledge, we further group all sessions. In the whole training process, PKIV-2 maintains an ensemble of projectors, which is significantly less than PKI. Similarly to PKIV-1, we {{initialize new weight $\bl{w}_p^{(t)}$ randomly}}, calculate the intermediate features $\{\bl{f}_i\}^{|\mc{D}^{(t)}|}$ based on dataset $\mc{D}^{(t)}$, and send them and the memory $\mc{M}^{(t)}$ to the ensemble. As shown in Figure~\ref{fig:vframework}, the ensemble takes $k$ as the baseline and divides the saved projector weights in previous sessions into $n$ groups, where $n = t//k$ is an integer. After input the feature $\bl{f}_i$, the output $\bl{v}$ is calculated as 
\begin{equation}
\begin{aligned}
\bl{v}_i = \phi_{p}(\bl{f}_i;\sum_{j=nk}^{t}\bl{w}_p^{(j)}) +\sum_{l=0}^{n}\phi_{p}(\bl{f}_i;\sum_{j=lk}^{lk+k}\bl{w}_p^{(j)}).
\end{aligned}
\label{eq:pkiv-2}
\end{equation}
In Eq.~\eqref{eq:pkiv-2}, PKIV-2 makes the resource consumption of the model $k$ times that of PKI, which greatly reduces the resources required by the model.

\section{Experiments}
To verify the effectiveness of our approach, we conduct experiments on three benchmarks and compare them with state-of-the-art methods.

\subsection{Experimental Setup}

\myPara{Datasets.} We conduct experiments on three datasets: CIFAR100~\citep{2009Learning}, MiniImageNet~\citep{russakovsky2015ijcv} and CUB200~\citep{Wah2011TheCB}. CIFAR100 and MiniImageNet have 100 classes with each class containing 500 training images and 100 testing images. CUB200 is a dataset for fine-grained image classiﬁcation containing 11,788 images of 200 classes. Following previous works~\citep{tao2020cvpr,zhang2021cvpr}, we split the 100 classes into 60 classes for the base session and 40 classes for incremental sessions. Then the 40 classes are divided into 8 different sets with a \emph{$5$-way $5$-shot} setting. For CUB200, 100 classes are used in the base session, and the other 100 classes are assigned as 10 incremental sessions with the setting of \emph{$10$-way $5$-shot}. The class and example assignments are random.

\myPara{Network.} Following \citet{yang2023iclr}, we adopt ResNet~\citep{he2016cvpr} as our backbone, and a three-layer MLP block as the projector. And we use ResNet12 without pretraining for CIFAR100 and MiniImageNet, and ResNet18 pre-trained on ImageNet for CUB200.

\myPara{Implementation.} We use SGD with momentum as the optimizer and adopt a cosine annealing strategy for learning rate. We train models with a batch size of $512$ in all datasets. We train for $100$ epochs in the base session, and also $100$ to $200$ iterations in each incremental session. In the base session, we use an initial learning rate of $0.25$ for CIFAR100 and MiniImageNet, and $0.01$ for CUB200. In the incremental sessions, we use $0.25$ as the learning rate for CIFAR100 and MiniImageNet, and $0.01$ as the learning rate for CUB200.
We use the standard data pre-processing and augmentation schemes to input images, including random resizing, random flipping, color jittering, Mixup~\citep{zhang2018iclr}, and CutMix~\citep{yun2019iccv}. 
Following the previous settings, the input images are rescaled into $32\times32$ for CIFAR100, $84\times84$ for MiniImageNet, and $224\times224$ for CUB200, respectively.

\begin{table}[t]
\centering
\resizebox{1.0\columnwidth}{!}{
\begin{tabular}{cccccccccccc}
\hline
	\multirow{2}{*}{Method} & \multicolumn{9}{c}{Accuracy in each session($\%$)$\uparrow$} & Average & Average \\
    &  0 & 1 & 2 & 3 & 4 & 5 & 6 & 7 & 8 & accuracy & improvement\\
	\hline
D-Cosine \citep{vinyals2016nips} & 74.55 & 67.43 & 63.63 & 59.55 & 56.11 & 53.80 & 51.68 & 49.67 & 47.68 & 58.23 & +10.29 \\	
 iCaRl \citep{rebuffi2017cvpr}  & 64.10 & 53.28 & 41.69 & 34.13 & 27.93 & 25.06 & 20.41 & 15.48 & 13.73 & 32.87 & +35.65  \\
        NCM \citep{hou2019cvpr} & 64.10 & 53.05 & 43.96 & 36.97 & 31.61  & 26.73 & 21.23 & 16.78 & 13.54 & 34.22 & +34.30  \\
                \hline

        TOPIC \citep{tao2020cvpr} & 64.10 & 55.88 & 47.07 & 45.16 & 40.11 & 36.38 & 33.96 & 31.55 & 29.37 & 42.62 & +25.90  \\
        SPPR \citep{zhu2021cvpr} & 64.10 & 65.86 & 61.36 & 57.45 & 53.69 & 50.75 & 48.58 & 45.66 & 43.25 & 54.52 & +14.00  \\
        CEC \citep{zhang2021cvpr} & 73.07 & 68.88 & 65.26 & 61.19 & 58.09 & 55.57 & 53.22 & 51.34 & 49.14 & 59.53 & +8.99  \\         
        LIMIT \citep{zhou2022tpami} & 73.81 & 72.09 & 67.87 & 63.89 & 60.70 & 57.77 & 55.67 & 53.52 & 51.23 & 61.84 & +6.68  \\
        MetaFSCIL \citep{chi2022cvpr} & 74.50 & 70.10 & 66.84 & 62.77 & 59.48 & 56.52 & 54.36 & 52.56 & 49.97 & 60.79 & +7.73 \\
        C-FSCIL \citep{hersche2022cvpr} & 77.47 & 72.40 & 67.47 & 63.25 & 59.84 & 56.95 & 54.42 & 52.47 & 50.47 & 61.64 & +6.88  \\
        Data-free Replay \citep{liu2022eccv} & 74.40 & 70.20 & 66.54 & 62.51 & 59.71 & 56.58 & 54.52 & 52.39 & 50.14 & 60.78 & +7.74 \\
        ALICE \citep{peng2022eccv} & 79.00 & 70.50 & 67.10 & 63.40 & 61.20 & 59.20 & 58.10 & 56.30 & 54.10 & 63.21 & +5.31 \\
        MCNet~\citep{ji2023tip} &73.30&69.34 &65.72&61.70&58.75 &56.44 &54.59 &53.01 &50.72 &60.40 &+7.78 \\
        DSN \citep{yang2023tpami} & 73.00 & 68.83 & 64.82 & 62.64 & 59.36 & 56.96 & 54.04 & 51.57 & 50.00 & 60.14 & +8.04  \\
        NC-FSCIL \citep{yang2023iclr} & \underline{82.52} & 76.82 & 73.34 & 69.68 & 66.19 & 62.85 & 60.96 & 59.02 & 56.11 & 67.50 & +1.02 \\
        CABD~\citep{zhao2023cvpr}	&79.45 	&75.38 	&71.84 	&67.95 	&64.96 	&61.95 	&60.16 	&57.67 	&55.88 	&66.14 	&+2.38 \\
        OrCo~\citep{ahmed2024cvpr}	&80.08 	&68.16 	&66.99 	&60.97 	&59.78 	&58.60 	&57.04 	&55.13 	&52.19 	&62.10 	&+6.42 \\
        OSHHG \citep{cui2024tmm} &63.55&62.88 &61.05 &58.13 &55.68 &54.59 &52.93 &50.39 &49.48 &56.52 &+12.00 \\
         EHS \citep{deng2024wacv} &71.27&67.40 &63.87 &60.40 &57.84 &55.09 &53.10 &51.45 &49.43 &58.87 &+9.65 \\
         \hline
    
	 \textbf{PKI} & \multirow{3}{*}{\textbf{83.02}} & \textbf{79.87} & \textbf{75.68} & \textbf{70.32} & \textbf{67.73} & \textbf{63.96} & \textbf{61.61} & \textbf{59.68} & \textbf{56.89} & \textbf{68.75} & -0.23 \\
  \textbf{PKIV-1} & & \underline{79.60} & \underline{75.21} & \underline{69.73} & 66.86 & 63.49 & \textbf{61.01} & 59.26 & 56.64 & 68.31 & +0.21 \\
  \textbf{PKIV-2} & & \underline{79.60} & \underline{75.21} & \underline{69.73} & \underline{67.46} & \underline{63.77} & \underline{61.53} & \underline{59.61} & \underline{56.74} & \underline{68.52} & - \\
\hline
\end{tabular}
}
\caption{FSCIL performance on CIFAR100. ``Average accuracy'' means the average accuracy of all sessions and ``Average improvement'' calculates the improvement of our approach over other methods. These methods include class-incremental learning or few-shot learning methods with FSCIL setting and FSCIL methods. The best and second-best results are in bold and underlined, respectively.}
\label{table1}
\end{table}

\begin{table}[ht]
\centering
\resizebox{1.0\columnwidth}{!}{
\begin{tabular}{cccccccccccc}
\hline
	\multirow{2}{*}{Method} & \multicolumn{9}{c}{Accuracy in each session($\%$)$\uparrow$} & Average & Average \\
    &  0 & 1 & 2 & 3 & 4 & 5 & 6 & 7 & 8 & accuracy & improvement\\
	\hline
	D-Cosine \citep{vinyals2016nips}& 70.37 & 65.45 & 61.41 & 58.00 & 54.81 & 51.89 & 49.10 & 47.27 & 45.63 & 55.99 & +12.51\\
	iCaRl \citep{rebuffi2017cvpr} & 61.31 & 46.32 & 42.94 & 37.63 & 30.49 & 24.00 & 20.89 & 18.80 & 17.21 & 33.29 & +35.22  \\
        NCM \citep{hou2019cvpr} & 61.31 & 47.80 & 39.30 & 31.90 & 25.70 & 21.40 & 18.70 & 17.20 & 14.17 & 30.83 & +37.68 \\
        \hline
        TOPIC \citep{tao2020cvpr} & 61.31 & 50.09 & 45.17 & 41.16 & 37.48 & 35.52 & 32.19 & 29.46 & 24.42 & 39.64 & +28.86  \\
        IDLVQ \citep{chen2021iclr} & 64.77 & 59.87 & 55.93 & 52.62 & 49.88 & 47.55 & 44.83 & 43.14 & 41.84 & 51.16 & +17.35 \\
        SPPR \citep{zhu2021cvpr} & 61.45 & 63.80 & 59.53 & 55.53 & 52.50 & 49.60 & 46.69 & 43.79 & 41.92 & 52.76 & +15.75  \\
        CEC \citep{zhang2021cvpr} & 72.00 & 66.83 & 62.97 & 59.43 & 56.70 & 53.73 & 51.19 & 49.24 & 47.63 & 57.75 & +10.76  \\
        LIMIT \citep{zhou2022tpami} & 72.32 & 68.47 & 64.30 & 60.78 & 57.95 & 55.07 & 52.70 & 50.72 & 49.19 & 59.06 & +9.45  \\
        Regularizer \citep{akyurek2022iclr} & 80.37 & 74.68 & 69.39 & 65.51 & 62.38 & 59.03 & 56.36 & 53.95 & 51.73 & 63.71 & +4.80 \\
        MetaFSCIL \citep{chi2022cvpr} & 72.04 & 67.94 & 63.77 & 60.29 & 57.58 & 55.16 & 52.90 & 50.79 & 49.19 & 58.85 & +9.66 \\
        C-FSCIL \citep{hersche2022cvpr} & 76.40 & 71.14 & 66.46 & 63.29 & 60.42 & 57.46 & 54.78 & 53.11 & 51.41 & 61.61 & +	6.90 \\
        Data-free Replay \citep{liu2022eccv} & 71.84 & 67.12 & 63.21 & 59.77 & 57.01 & 53.95 & 51.55 & 49.52 & 48.21 & 58.02 & +10.49\\
        ALICE \citep{peng2022eccv} & 80.60 & 70.60 & 67.40 & 64.50 & 62.50 & 60.00 & 57.80 & 56.80 & 55.70 & 63.99 & +4.52 \\
        MCNet~\citep{ji2023tip}&72.33&67.70 &63.50 &60.34 &57.59 &54.70 &52.13 &50.41 &49.08 &58.64 &+9.87 \\
        NC-FSCIL \citep{yang2023iclr} & \underline{84.02} & 76.80 & 72.00 & 67.83 & 66.35 & 64.04 & 61.46 & 59.54 & 58.31 & 67.82 & +	0.69 \\
        CABD~\citep{zhao2023cvpr}	&74.65 	&70.43 	&66.29 	&62.77 	&60.75 	&57.24 	&54.79 	&53.65 	&52.22 	&61.42 	&+7.09 \\
        OrCo~\citep{ahmed2024cvpr}	&83.30 	&75.32 	&71.53 	&68.16 &	65.63 	&63.12 	&60.20 	&58.82 	&58.08 	&67.13 	&+1.38 \\
        OSHHG \citep{cui2024tmm}&60.65&59.00 &56.59 &54.78 &53.02 &50.73 &48.46 &47.34 &46.75 &53.04 &+15.47\\
        EHS \citep{deng2024wacv} &69.43&64.86 &61.30 &58.21 	&55.49 	&52.77&50.22 &48.61 &47.67 &56.51 &+12.00\\
        \hline
         \textbf{PKI} & \multirow{3}{*}{\textbf{84.38}}  & \textbf{78.71} & \textbf{74.20 } & \textbf{68.33}  & \textbf{67.29} & \textbf{64.91} & \textbf{61.97} & \textbf{60.06} & \textbf{58.85} & \textbf{68.74} & -0.24 \\
         \textbf{PKIV-1} && \underline{78.34} & \underline{73.73} & \underline{67.92}  & 66.53 & 64.54 & 61.48 & 59.64 & 58.40 & 68.33 & +0.18 \\
         \textbf{PKIV-2} &  & \underline{78.34} & \underline{73.73} & \underline{67.92}  & \underline{66.88} & \underline{64.82} & \underline{61.76} & \underline{59.99} & \underline{58.74} & \underline{68.51} & - \\

\hline
\end{tabular}
}
\caption{FSCIL performance on MiniImageNet. ``Average accuracy'' means the average accuracy of all sessions and ``Average improvement'' calculates the improvement of our approach over other methods. These methods include class-incremental learning or few-shot learning methods with FSCIL setting and FSCIL methods. The best and second-best results are in bold and underlined, respectively.}
\label{table2}
\end{table}

\begin{table*}[ht]
\centering
\resizebox{1.0\columnwidth}{!}{
\begin{tabular}{cccccccccccccc}
\hline
	\multirow{2}{*}{Method} & \multicolumn{11}{c}{Accuracy in each session($\%$)$\uparrow$} & Avg. & Avg. \\
    &  0 & 1 & 2 & 3 & 4 & 5 & 6 & 7 & 8 & 9 & 10 & acc. & imp.\\
	\hline
	D-Cosine \citep{vinyals2016nips} & 75.52 & 70.95 & 66.46 & 61.20 & 60.86 & 56.88 & 55.40 & 53.49 & 51.94 & 50.93 & 49.31 & 59.36 & +8.38 \\
        iCaRl \citep{rebuffi2017cvpr} & 68.68 & 52.65 & 48.61 & 44.16 & 36.62 & 29.52 & 27.83 & 26.26 & 24.01 & 23.89 & 21.16 & 36.67 & +31.06   \\
        NCM \citep{hou2019cvpr} & 68.68 & 57.12 & 44.21 & 28.78 & 26.71 & 25.66 & 24.62 & 21.52 & 20.12 & 20.06 & 19.87 & 32.49 & +35.25\\
       \hline
        TOPIC \citep{tao2020cvpr} & 68.68 & 62.49 & 54.81 & 49.99 & 45.25 & 41.40 & 38.35 & 35.36 & 32.22 & 28.31 & 26.28 & 43.92 & +23.82 \\
        IDLVQ \citep{chen2021iclr} & 77.37 & 74.72 & 70.28 & 67.13 & 65.34 & 63.52 & 62.10 & 61.54 & 59.04 & 58.68 & 57.81 & 65.23 & +2.51\\
        RRFSP~\citep{cheraghian2021iccv}& 68.78 & 59.37 & 59.32 & 54.96 & 52.58 & 49.81 & 48.09 & 46.32 & 44.33 & 43.43 & 43.23 & 51.84 & +15.90\\
        SPPR \citep{zhu2021cvpr} & 68.68 & 61.85 & 57.43 & 52.68 & 50.19 & 46.88 & 44.65 & 43.07 & 40.17 & 39.63 & 37.33 & 49.32 & +18.42\\
        CEC \citep{zhang2021cvpr} & 75.85 & 71.94 & 68.50 & 63.50 & 62.43 & 58.27 & 57.73 & 55.81 & 54.83 & 53.52 & 52.28 & 61.33 & +6.81 \\ 	
        LIMIT\citep{zhou2022tpami}& 76.32 & 74.18 & 72.68 & 69.19 & 68.79 & 65.64 & 63.57 & 62.69 & 61.47 & 60.44 & 58.45 & 66.67 & +0.97\\
        MetaFSCIL \citep{chi2022cvpr} & 75.9 & 72.41 & 68.78 & 64.78 & 62.96 & 59.99 & 58.3 & 56.85 & 54.78 & 53.82 & 52.64 & 61.93 & +5.81\\
        FACT \citep{zhou2022cvpr} & 75.90 & 73.23 & 70.84 & 66.13 & 65.56 & 62.15 & 61.74 & 59.83 & 58.41 & 57.89 & 56.94 & 64.42 & +3.32\\
        Data-free replay \citep{liu2022eccv} & 75.90 & 72.14 & 68.64  &63.76 & 62.58 & 59.11 & 57.82 & 55.89 & 54.92 & 53.58 & 52.39 & 61.52 & +6.22\\
        ALICE \citep{peng2022eccv} & 77.40 & 72.70 & 70.60 & 67.20 & 65.90 & 63.40 & 62.90 & 61.90 & 60.50 & \underline{60.60} & 60.10 & 65.75 & +1.99\\
        MCNet~\citep{ji2023tip} &77.57 &73.96 &70.47 &65.81 &66.16 &63.81 &62.09 &61.82 &60.41 &60.09 &59.08 &65.57  &+2.17 \\
      NC-FSCIL \citep{yang2023iclr} & \underline{80.45} & {75.98} & {72.30} & {70.28} & {68.17} & {65.16} & 64.43 & 63.25 & {60.66} & {60.01} & {59.44} & {67.28} & +0.46\\
      CABD~\citep{zhao2023cvpr}&79.12 	&75.37 &	72.80 	&69.05 	&67.53 	&65.12 	&64.00 	&63.51 	&\textbf{61.87} 	&\textbf{61.47} 	&\textbf{60.93}	&67.34 	&+0.40 \\
    OrCo~\citep{zhao2023cvpr}	&75.59 	&66.85 	&64.05 	&63.69 	&62.20 	&60.38 	&60.18 	&59.20 	&58.00 	&58.42 	&57.94 &	62.41 	&+5.33 \\
      MgSvF \citep{zhao2024tpami} & 72.29 & 70.53 & 67.00 & 64.92 & 62.67 & 61.89 & 59.63 & 59.15 & 57.73 & 55.92 & 54.33 & 62.37 & +5.37 \\
      OSHHG \citep{cui2024tmm} &63.20 &62.61 &59.83 &56.82 &55.07 &53.06 &51.56&50.05&47.50&46.82&45.87&53.85 &+13.88\\
      \hline
\textbf{PKI} & \multirow{3}{*}{\textbf{80.51}} & \textbf{77.25} & \textbf{73.98} & \textbf{70.69} & \textbf{68.93} & \textbf{66.07} & \textbf{64.89} & \textbf{63.91} & \underline{61.01} & 60.46 & \underline{59.93} & \textbf{67.97} & -0.23 \\
 \textbf{PKIV-1} & & \underline{76.61} & \underline{73.72} & \underline{70.36} & {68.54} & 65.61 & {64.48} & {63.49} & {60.31} & 59.83 & {59.22} & 67.52 & +0.22\\
 \textbf{PKIV-2} & & \underline{76.61} & \underline{73.72} & \underline{70.36} & \underline{68.81} & \underline{65.74} & \underline{64.78} & \underline{63.85} & {60.89} & 60.14 & {59.68} & \underline{67.74} & - \\
\hline
\end{tabular}
}
\caption{FSCIL performance on CUB200. ``Avg. acc.'' means the average accuracy of all sessions and ``Avg. imp.'' calculates the improvement of our approach over other methods. These methods include class-incremental learning or few-shot learning methods with FSCIL setting and FSCIL methods. The best and second-best results are in bold and underlined, respectively.}
\label{table3}
\end{table*}

\subsection{State-of-the-art Comparison}


We conduct experimental comparisons with state-of-the-art on CIFAR100, MiniImageNet, and CUB200. The results are reported in Table~\ref{table1}, Table~\ref{table2} and Table~\ref{table3}, respectively. We compare two groups of methods: class-incremental learning or few-shot learning methods under the FSCIL setting, and the FSCIL methods. We find several meaningful conclusions based on the PKIV-2 from the results. First, our approach achieves the best average accuracy of $ 68.52\% $ on CIFAR100, $68.51\%$ on MiniImageNet, and $67.84\%$ on CUB200. Compared with the recent advanced approach NC-FSCIL~\citep{yang2023iclr}, our PKI has an accuracy improvement of $1.02\%$ on CIFAR100, $0.69\%$ on MiniImageNet and $0.46\%$ on CUB-200, respectively. These results show that our approach maintains outstanding performance. Second, our approach achieves the best performance in all sessions on both CIFAR100 and MiniImageNet, which is $0.15\%-2.78\%$ on CIFAR100 and $0.09\%-1.54\%$ on MiniImageNet higher than NC-FSCIL in all sessions. The main reason is the effective utilization of rich prior knowledge greatly alleviates catastrophic forgetting. Third, our approach delivers the best average accuracy on CUB200 but a slightly lower accuracy in the last few sessions. We argue high similarity between old and new classes will challenge our discriminability with a few examples.

\begin{table*}[ht]
\centering
\resizebox{0.7\columnwidth}{!}{
\begin{tabular}{ccccccccccc}
\hline
\multirow{2}{*}{$k$} &  \multicolumn{9}{c}{Accuracy in each session($\%$)$\uparrow$}  \\
    & 0 & 1 & 2 & 3 & 4 & 5 & 6 & 7 & 8\\
	\hline
1  (PKI)	&\multirow{5}{*}{83.02} & 	79.87 & 	75.68 & 	70.32 & 	67.73 & 	63.96 & 	61.61 & 	59.68 & 	56.89 \\
2	& & 	79.60 & 	75.21 & 	69.86 & 	67.54 & 	63.78 & 	61.62 & 	59.58 & 	56.73 \\
3  (PKIV-2)& & 	79.60 & 	75.21 & 	69.83 & 	67.46 & 	63.77 & 	61.53 & 	59.61 & 	56.74 \\
4	& & 	79.60 & 	75.21 & 	69.83 & 	66.86 & 	63.64 & 	61.26 & 	59.32 & 	56.66 \\
T  (PKIV-1)	& & 	79.60 & 	75.21 & 	69.83 & 	66.86 & 	63.49 & 	61.01 & 	59.26 & 	56.64 \\
\hline
\end{tabular}
}
\caption{The effect of $k$. We report the accuracy on CIFAR100.}
\label{ep}
\end{table*}


\subsection{Ablation Study}

\myPara{Effect of network structure and $k$.} As shown in Table~\ref{table1}, Table~\ref{table2} and Table~\ref{table3}, PKI leads the accuracy evaluation in all sessions compared to the two modified variants PKIV-1 and PKIV-2. This significant advantage shows that PKI performs well in terms of both accuracy and reliability. However, while pursuing high accuracy, PKI may require more resources to support its operation, which may become a disadvantage in some resource-constrained scenarios. In contrast, PKIV-2 finds a balance between performance and resource consumption. Although the accuracy of PKIV-2 is slightly inferior to that of PKI, it does not decrease significantly and maintains a high level. Notably, PKIV-2 is significantly lower than PKI in resource consumption, which is suitable for a wider range of environments while maintaining high performance. Compared with other state-of-the-art methods, PKIV-1 also achieves high accuracy while consuming few resources. As shown in Table~\ref{ep}, the performance gradually improves as the $k$ decreases. Specifically, after constructing a new projector, there is a notable improvement observed, indicating that the addition of projectors plays a positive role in promoting the improvement of model processing power and efficiency. However, the number of projectors can not be increased indefinitely. A large number of projectors not only impose additional computational and storage burdens on the model but also complicate the calculations, which can adversely affect the performance. Therefore, we choose $k=3$ to strike a good balance between performance and resource consumption. This choice ensures sufficient performance gains but also avoids the waste of resources and performance bottlenecks caused by too many projectors. 

\begin{figure*}[t]
\centering
\includegraphics[width=0.9\linewidth]{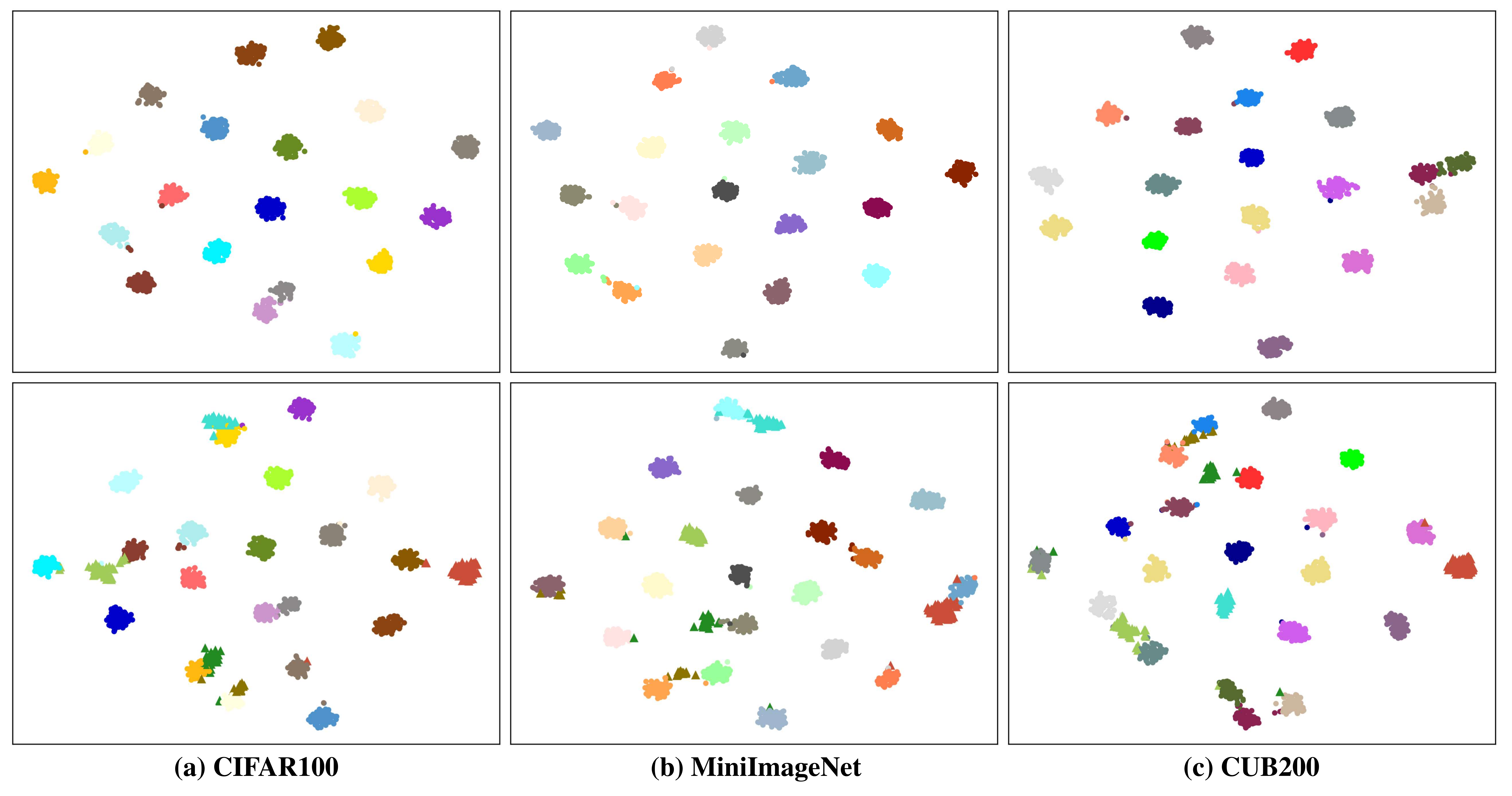}
\caption{Representation visualization with t-SNE~\citep{maaten2008jmlr}, which uses the base session and the last session as an example. We randomly select $50$ examples over $20$ base classes and $5$ incremental classes to show the model effect. Symbols `$\bullet$' and `$\blacktriangle$' represent examples of base classes and incremental classes, respectively. (a), (b) and (c) are visual features on CIFAR100, MiniImageNet, and CUB200, respectively. The top row is visual features in the base session, and the bottom row is visual features in the last incremental session.}
\label{fig:tsne}
\end{figure*}

\begin{wraptable}[9]{r}{0.4\linewidth} 
\vspace{-1.0em}
\begin{center}
\resizebox{0.4\columnwidth}{!}{
\begin{tabular}{cccc}
\hline
\multirow{2}{*}{number}  &\multicolumn{3}{c}{Accuracy in sessions ($\%$)}\\
    & 0 & 1 & 8\\
\hline
1&82.62&	78.32&	53.58\\
2&82.93&	79.14&	56.18\\
3&83.02& 	\textbf{79.60}& 	\textbf{56.74} \\
4&\textbf{83.14}& 	79.35&	55.36\\
\hline
\end{tabular}
}
\caption{The effect of layer number of the projector on CIFAR100.}
\label{tab:layer}
\end{center}
\end{wraptable}
\myPara{Effect of the layer number of the projector.} We further check the effect of the layer number in each projector as shown in Table \ref{tab:layer}. The accuracy of the model shows an upward trend as the number of projectors is increased, which means more layers can give the model stronger nonlinear fitting ability and 
enable it to capture more complex and detailed features. However, limited data may not be enough to support finetuning too many layers in incremental sessions, which leads to overfitting or inefficient training. The results reveal that the projector with a certain layer number makes a good feature-classifier alignment to balance old-class memorizing and novel-class learning. Therefore, we use a three-layer MLP as our projector, which ensures that the model has sufficient complexity to capture the key information and avoid poor results. 

\subsection{Further Discussion}

\myPara{Representation visualization.} To show the classification ability of our approach, we randomly select the examples in several base classes and incremental classes on three datasets, and then visualize their representations with t-SNE~\citep{maaten2008jmlr}. As shown in Figure~\ref{fig:tsne}, we can find that the examples in the same classes can be clustered well in the base session. In the last incremental session, although the feature maps of some old classes are shifted, most classes are still clustered in the feature space.

\begin{wrapfigure}[13]{r}{0.5\linewidth} 
\centering
\includegraphics[width=0.85\linewidth]{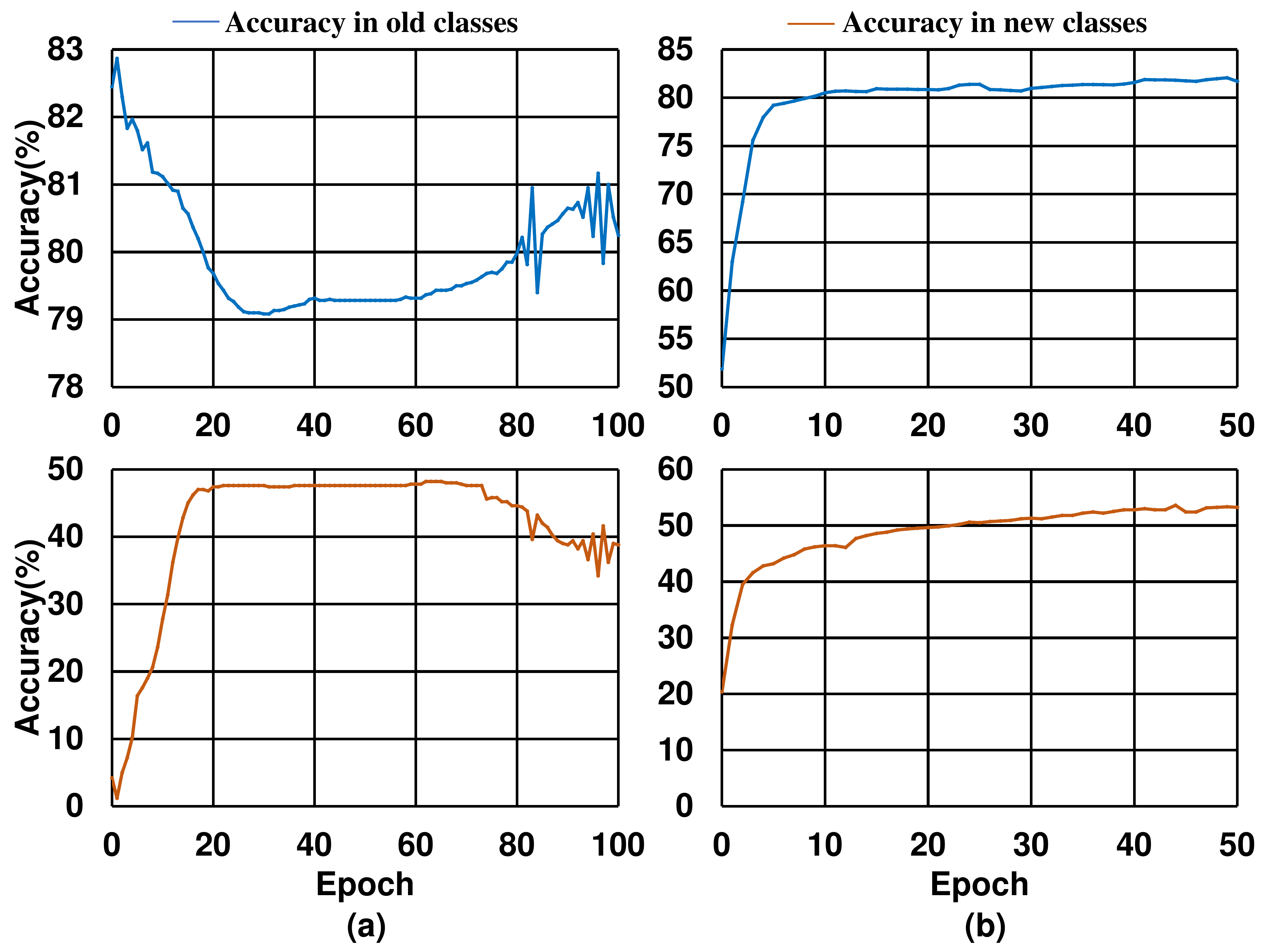}
\caption{{{Projector initialization in session $1$ on CIFAR100. (a) and (b) initialize the current projector with the projector from the previous session, and in random, respectively.}}}
\label{fig:init}
\end{wrapfigure}

\myPara{{{Projector initialization and capability.}}}~{{We first check the effect of projector initialization. From Figure~\ref{fig:init} (a), when initialing the new projector }}
{{with the previous projector weights, the performance of old classes initially declines and then tends to increase, while the model adapts to new classes in initial rounds and then overfit. From Figure~\ref{fig:init} (b), when initialing the new projector randomly, the model adapts to new classes while recalling old classes, achieving higher performance. Therefore, the projectors start from the same initialization can achieve better performance.}}

\begin{figure}[t]
	\centering
	\includegraphics[width=1.0\linewidth]{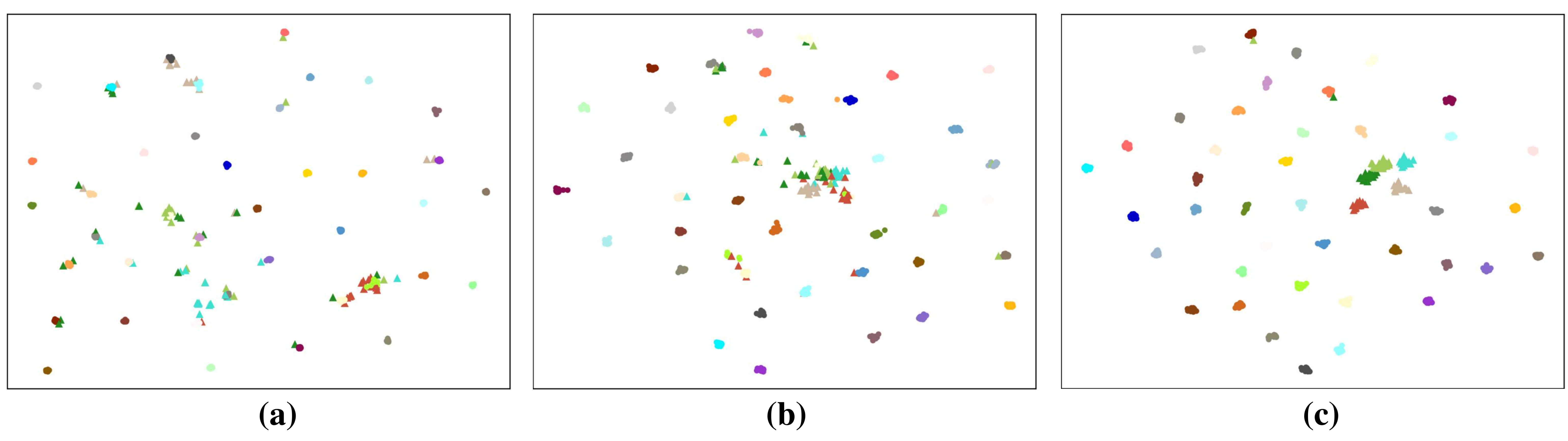}
	\caption{{{Visualization on projector capability of session $1$ on CIFAR100. We randomly select 20 examples over 40 base classes and 5 incremental classes. Symbols `$\bullet$' and `$\blacktriangle$' represent examples of base classes and incremental classes, respectively. (a) and (b) are visual features generated only using previous projectors and only the current projector, respectively. (c) is visual features of all classes using all projectors.}}}
	\label{fig:cap}
\end{figure}

{{To study the capability of different projectors, Figure~\ref{fig:cap} visualizes their roles. Previous projectors distinguish old classes well, while the new projector learns new knowledge well and supplements old knowledge. And the collaboration between the old and new projectors distinguishes all samples well, which implies the effectiveness of the ensemble design. Table~\ref{weight} further shows the effect of the weighting factor $\alpha$ in adjusting the influence of new projectors. With $\alpha$ increases, the ability of the model to identify new classes gradually increases, whereas a larger $\alpha$ expands the adjustable range of features. Additionally, compared with PKIV-2, PKIV-1 achieves a more stable balance by sharing the same projector across sessions.}}



\begin{table}[t]
\centering
\resizebox{0.75\columnwidth}{!}{
\begin{tabular}{c|c|cc|cc|cc}
\hline
\multirow{2}{*}{~{{$\alpha$}}~ }& \multirow{2}{*}{~{{Base}}~}& \multicolumn{2}{c}{~~{PKI}~~}&\multicolumn{2}{c}{~~{PKIV-1}~~}&\multicolumn{2}{c}{~~{PKIV-2}~~}\\
     &  & ~~~{First}~ & ~{Final}~~~  & ~~~{First}~ & ~{Final}~~~ &  ~~~{First}~ & ~{Final}~~~\\
	\hline
{0.2} &	{\multirow{5}{*}{83.02}}&{77.21} & {52.74}	&{77.12} & {52.36} 	&{77.12} & {52.13}\\
{0.4} &	&{78.46} & {54.36} &{78.03} &{53.80} &{78.03} &{53.64}\\
{0.6} &	& {79.06} & {55.42 }& {78.61} &{54.67}&{78.61 }&{54.51}\\
{0.8} &	&{79.75} & {56.69} & {79.34} &{55.87} &{79.34 }&{55.93}\\
{1.0} &   &{79.87} & {56.89} &{79.60} &{56.64} &{79.60} &{56.74}\\
\hline
\end{tabular}
}
\caption{{{We introduce a factor $\alpha$ to adjust the influence of new projectors constructed in incremental sessions on CIFAR100. In PKI, we apply $\alpha$ to all features obtained from new projectors in incremental sessions. For two variants, we apply $\alpha$ to all projector weights constructed in incremental sessions. ``Base'' denotes the accuracy of the base session. ``First'' and ``Final'' denote the accuracy of the first and last incremental sessions.}}}
\label{weight}
\end{table}


\myPara{Catastrophic forgetting.}~It is a challenging problem in incremental learning and one key is to make the model retain memory like humans. PKI and two variants mitigate catastrophic forgetting by retaining more prior knowledge during incremental learning. We freeze the backbone to utilize the previously learned feature extraction capabilities and reuse previous projector weights, which can retain old knowledge as shown in Table~\ref{table1}, Table~\ref{table2}, Table~\ref{table3}, and Table~\ref{ep}. In this way, our approach mitigates catastrophic forgetting and improves performance, resulting in a more stable and sustainable learning process.

\myPara{Overfitting.}~In all incremental sessions, it is easy to overfit to novel classes due to the examples of new classes are not sufficient. We use new projectors to learn new knowledge in incremental sessions, which can be quickly fitted based on prior knowledge. In this way, our approach can use fewer training rounds and fewer parameters to learn models without overfitting. 

\myPara{{{Prior knowledge stored in PKI.}}}~{{Our PKI stores sufficient prior knowledge across several key components. First, the frozen backbone preserves some common knowledge from the base session and mitigates overfitting. Second, the memory retains a wealth of old knowledge accurately and steadily, which indirectly helps the model to distinguish new and old classes more accurately. Finally, the frozen projectors or parameters from previous sessions store more old knowledge, which is a foundation for subsequent sessions. Prior knowledge plays an important role in PKI, effectively alleviating catastrophic forgetting and overfitting.}}

\myPara{{{Relationship to other methods.}}}~{{Unlike previous methods, our method emphasizes the importance of prior knowledge and integrates it by leveraging additional networks. Beyond the memory and the frozen backbone, we retain additional networks to preserve richer prior knowledge, establishing a stronger foundation for subsequent sessions. We also cascade an ensemble of projectors to effectively integrate and utilize prior knowledge, enhancing the ability to distinguish between old and new classes. Further, we design two variants to achieve an optimal balance between accuracy and resources.}}

\myPara{Resource consumption.} {{Similar to recent methods, we build a memory $\mathcal{M}$ to retain more old knowledge, which stores $\cup_{i=0}^{t-1} L^{(i)}$ features in the current session (a feature per class). Compared with other storage requirements, the storage consumed by the memory $\mathcal{M}$ is negligible ($0.24$MB to store the memory on CIFAR100).}} Different from recent methods, our approach builds extra projectors to learn old and new knowledge. On the one hand, the new projectors capture specific features or patterns in new data without interfering with old knowledge, reducing the learning rounds and time. Even though more operations are included in the training process, the cost of time for each round is not obvious due to the small number of samples. On the other hand, storage space gradually accumulates and memory consumption gradually increases as the number of sessions increases. In the base session, we use only one projector for model training, which consumes $p$ parameters. In incremental sessions, we consume $Tp$ parameters for PKI, $2p$ parameters for PKIV-1 and $np$ parameters for PKIV-2. Compared with other storage requirements, such as backbone networks with nearly 12M parameters, the ensemble of projectors costs only a relatively small amount of parameters.  



\section{Conclusion}
In this work, we introduce a prior knowledge-infused neural network (PKI) to enhance the performance of FSCIL by tackling inherent challenges: catastrophic forgetting and overfitting. PKI keeps rich prior knowledge incrementally for current and subsequent sessions by cascading an ensemble of projectors and freezing parameters. The prior knowledge not only provides additional regularization information but also accelerates the model training. By building a new projector and maintaining previous projectors, the network enables efficient feature-classifier alignment for all classes, helping the model mitigate catastrophic forgetting and learn new concepts quickly. Additionally, the risk of overfitting is overcome by fewer parameter updating and regularization constraints. To further decrease resource consumption associated with the projectors, we propose two variants of the PKI (PKIV-1 and PKIV-2), which involve reducing the number of projectors. Extensive experiments are conducted to demonstrate the effectiveness of our approach. In the future, we will optimize the efficiency of the approach and extend its application in practical scenarios. 

\bibliographystyle{elsarticle-harv}
\bibliography{main}

\end{document}